\documentclass{article}

\pdfoutput=1 
\PassOptionsToPackage{numbers}{natbib}
\usepackage[preprint]{neurips_2020}
\usepackage[utf8]{inputenc} 
\usepackage[T1]{fontenc}    
\usepackage{hyperref}       
\usepackage{url}            
\usepackage{booktabs}       
\usepackage{amsfonts}       
\usepackage{nicefrac}       
\usepackage{microtype}      

\usepackage{booktabs} 
\usepackage{amssymb}
\usepackage{amsmath}
\usepackage{xcolor}
\usepackage{graphicx}
\usepackage{float}

\graphicspath{ {./images/} }

\usepackage[export]{adjustbox}
\usepackage{subfig}
\usepackage{wrapfig}
\usepackage{threeparttable}
\usepackage{enumitem}
\usepackage{hyperref}

\newcommand{\q}[1]{"#1"}

\title{Hierarchical nucleation in deep neural networks}
\author{%
  Diego Doimo \\
International School for Advanced Studies \\
  \texttt{ddoimo@sissa.it}\\  
   \And
   Aldo Glielmo \\
International School for Advanced Studies \\
   \texttt{aglielmo@sissa.it} \\
   \And
   Alessio Ansuini \\
Area Science Park\\
   \texttt{alessio.ansuini@areasciencepark.it} \\
      \And
   Alessandro Laio \\
International School for Advanced Studies\\
   \texttt{laio@sissa.it} \\
}

\begin{document}

\maketitle

\begin{abstract}
Deep convolutional networks (DCNs) learn meaningful representations where data that share the same abstract characteristics are positioned  closer and  closer. Understanding these representations and how they are generated is of unquestioned practical and theoretical interest. In this work we study the evolution of the probability density of the ImageNet dataset across the hidden layers in some state-of-the-art DCNs.
We find that the initial layers generate a unimodal probability density getting rid of any structure irrelevant for classification. In subsequent layers density peaks arise in a hierarchical fashion that mirrors the semantic hierarchy of the concepts. Density peaks corresponding to single categories appear only close to the output and via a very sharp transition which resembles the nucleation process of a heterogeneous liquid. This process leaves a  footprint in the probability density of the output layer where the topography of the  peaks allows reconstructing the semantic relationships of the categories.
 \end{abstract}
 
\section{Introduction}

Deep convolutional networks (DCNs) have become fundamental tools of modern science and technology.
They provide a powerful approach to supervised classification, allowing the automatic extraction of meaningful features from data. 
The capability of DCNs to discover representations without  human input, has attracted the interest of the machine learning community.
In the intermediate layers of a DCN,  the data are represented with a set of features (the activations) embedded in  a manifold whose tangent directions capture the relevant factors of variation of the inputs \cite{bengio2013representation, goldt2020modelling}.
Accordingly, understanding these data representations requires both studying the geometrical properties of the underlying manifolds and characterising the data distributions on them.

In the present paper, we analyse how the probability density of the data on the supporting manifold changes across the layers of a DCN. We consider in particular DCNs trained for classifying ImageNet; as we will see, the complexity and heterogeneity of this dataset critically affects the results of our analysis. 

Comparisons between representations based on generalizations of multivariate correlation have already been performed with the methods in \cite{svcca} (SVCCA), \cite{morcos2018insights} (PWCCA) and, more extensively, in \cite{cka} (CKA).
Representational similarity analysis (RSA) \cite{kriegeskorte2008representational} -- introduced originally in neuroscience -- investigates artificial representations as well, and in each layer a matrix of pairwise distances (representation dissimilarity matrix (RDM) ) between the activation vectors of the data points tells which data is similar or dissimilar in that layer.
The introduction of RDMs allowed performing multiple comparisons including those between artificial and biological networks \cite{khaligh2014deep, yamins2014performance, cadieu2014deep}.

More recently, the question whether DCNs learn a hierarchy of classes was addressed, exploiting class confusion patterns \cite{bilal2017convolutional}.
Other studies investigated more specifically geometrical and structural properties of the representations. In \cite{id} a common trend in the intrinsic dimension was found across several architectures; in \cite{salakhutdinov2007learning} the soft-neighbor loss  was used as a tool to reveal structural changes in neighbors organization across layers, also during training \cite{frosst2019analyzing}.

\begin{figure}[!t]
\centering
{\includegraphics[width=0.8\columnwidth]{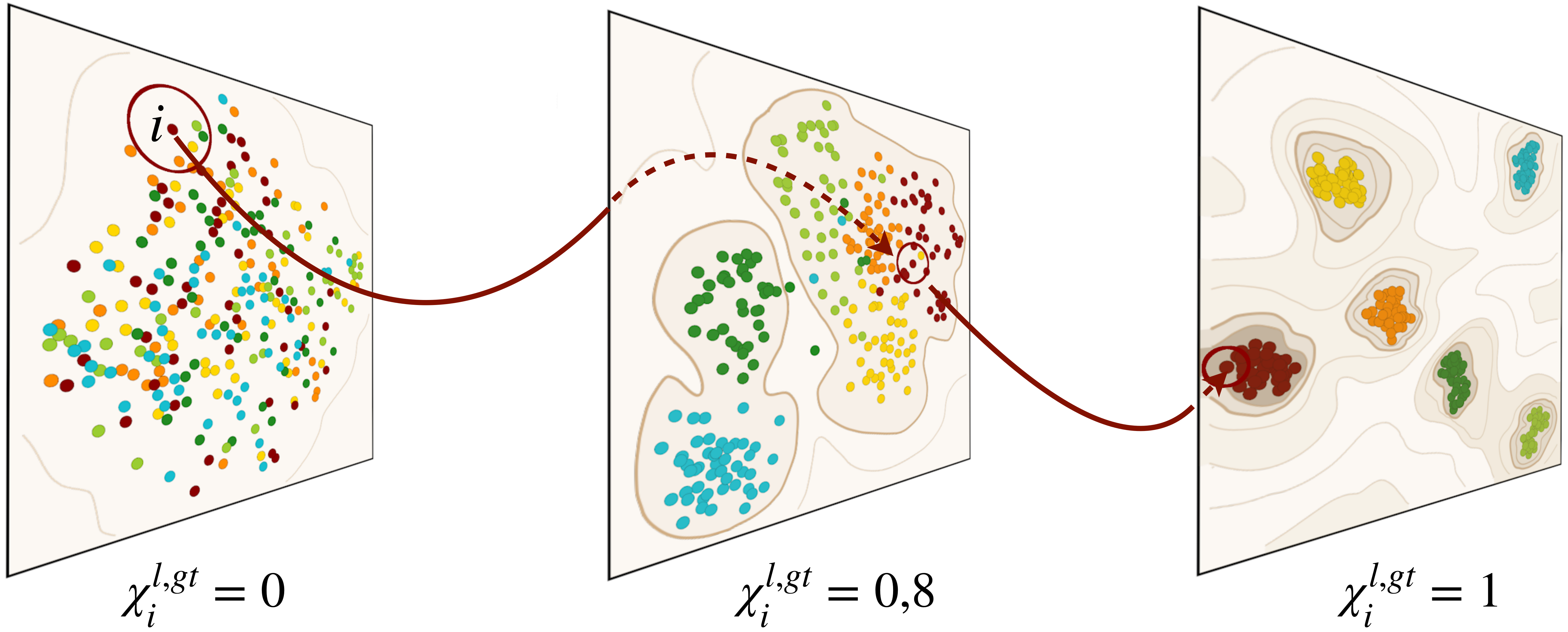} \label{fig:flow}} 
	\caption{{\bf Evolution of the data representations in ResNet152}. Projections of the representations of the input (left), conv4 (middle) and output  layers in ResNet152 for six ImageNet classes. Contours schematically  portray the density isolines on the data manifold. The dark red circles surround the five nearest neighbors of a point $i$; $\chi_i^{l, gt}$  represents the fraction of these points that are in the same class of $i$.}
\label{fig:cartoon}
\vspace{-0.5cm}
\end{figure}
Here we take a complementary perspective. One can view a DCN as an engine capable of iteratively shaping a probability density. The input data can be seen as instances harvested from a given probability distribution. 
This distribution is then modified again and again by applying, at each layer, a non-linear transformation to the data coordinates. The result of this sequence of transformations is well  understood:  in the output layer of  a trained network, data belonging to different categories form well separated clusters, which can be viewed as distinct peaks of a probability density (see Fig. \ref{fig:cartoon}).
But where in the network do these peaks appear? Do they develop slowly and gradually or all of a sudden? Is this change model-specific or is it shared across architectures? And what is the probability \emph{flux} between a layer and the next? 
In the input layer the data points are mixed: data with different ground truth labels are close to each other. In the output layer, the neighborhood of each data point is ordered, namely it contains mostly data points belonging to the same category. Where in the network does the transition from disordered  to ordered neighborhoods take place? 
Is it simultaneous with the formation of the probability peaks?

The pivotal role of depth in determining the accuracy of a neural network suggests that the transformation of the probability density should be slow and gradual in order to be effective. The analysis reported in \cite{svcca, morcos2018insights, cka} are consistent with this scenario. However, we will see that  especially in a DCN trained for a complex classification task the evolution of the probability density is not really smooth, with spikes in the probability flux and sudden changes in the modality.  

We analyse the probability landscape in the intermediate layers of a DCN 
by a technique which allows estimating the probability density and characterizing its features even if this is defined as a function of hundreds of thousands of coordinates, provided that the data are embedded in a relatively low dimensional manifold \cite{cluster}. The advantage of this approach with respect to standard dimensional reduction  techniques is that the embedding manifold does not necessarily have to be an hyperplane, but can be arbitrarily curved, twisted and topologically complex. To analyse the probability flux  between the layers we  use an extension of neighboring hit \cite{NH}.
The main results of this analysis also sketched in Fig. \ref{fig:cartoon} can be summarized as follows:

\begin{itemize}
    \item Representations in DCNs trained for complex classification tasks  do not evolve smoothly, but through nucleation-like events, in which the neighborhood of a data point changes rather suddenly (Sec. \ref{sec:nucl}); 
    
    \item In the first layers of the network any structure which is initially present in the probability density of the input is washed out, reaching a state  with a single probability peak where the neighborhoods mainly contain simple images characterized by elementary geometrical shapes (Sec. \ref{sec:entropy});
    
    \item In the successive layers a structure in the landscape starts to emerge, with probability peaks appearing in an order that mirrors the semantic hierarchy of the dataset: neighborhoods are first populated by images that share the same high-level attributes (Sec. \ref{sec:cluster});
    
    \item In the output layer the probability landscape is formed by density peaks containing data points with the same ground truth classification; interestingly, these peaks  are organized in complex "mountain chains" resembling the semantic kinship of the categories (Sec. \ref{sec:cluster}). 
    
\end{itemize}
It short, we find that the disorder-order transition induced by a trained DCN can be characterized without any reference to the ground truth categories as a sequence of changes in the modality of the  probability density of the representations.
These changes are achieved by reshuffling the neighbors of the data points again and again, in a process which resembles the diffusion in an heterogeneous liquid, followed by the nucleation of an ordered phase. 

\section{Methods}
The DCNs we consider in this work   are classifiers ${\bf y} = f({\bf x})$ that map a data point ${\bf x}_i \in \mathbb{R}^p$, for example an image, to its categorical target ${\bf y}_i \in \{0, 1\}^q $ typically encoded with a one-hot vector of dimension $q$ equal to the number of classes.
Feedforward networks achieve the task via a function composition  $f = f^{(1)} \to f^{(2)} \to ... \to f^{(L)}$ that transform the input sequentially ${\bf x}_i \to  {\bf x}^{(1)}_i \to ... \to  {\bf x}^{(L)}_i $.
We call the vector ${\bf x}^{(l)}_i$ containing the value of the activations of the $l$-th layer for data point $i$  the {\it representation} of ${\bf x}_i$ at the layer $l$. 
The sequence of representations of these datapoints on a trained network can be seen as a \q{trajectory} in a very high dimensional space.
The relative positions of the $N$ inputs change from an initial state where the neighborhood of each point contains members of different classes to a final state where images of the same class have been mapped close together to the same target point.
We study this process with two approaches, one aimed at describing the probability flux across the layers (Sec. \ref{sec:ov}) the other aimed at characterizing the features of the probability density in each layer (Sec. \ref{sec:density_peak}).

\subsection{The neighborhood overlap}\label{sec:ov}

Let $\mathcal{N}_{k}^l(i) $ be the set of $k$ points nearest to  ${\bf x}_i^{(l)} $ in euclidean distance at a given layer $l$, and let $A^l$ be an $N \times N$ adjacency matrix with entries 
$A^l_{{ij}} =1$ if $j \in \mathcal{N}_k^l(i) $ and $0$ otherwise.
Through $A$ we define an index of similarity $\chi_k^{l,m} \in [0, 1]$ between two layers $l$ and $m$  as: 
\begin{equation}
\chi_k^{l,m} = \dfrac{1}{N} \sum_i \dfrac{1}{k} \sum_j {A^l}_{ij} A^m_{ij}   
\label{eq:overlap}
\end{equation}
The similarity just introduced has a very intuitive interpretation: it is the average fraction of common neighbors in the two layers considered: for this reason, we will refer to $\chi_k^{l,m}$ as the \emph{neighborhood overlap} between layers $l$ and $m$.

In the same framework we also compare the similarity of a layer with the ground truth categorical classification  defining the \q{ground truth} adjacency matrix
$A^{gt}_{{ij}} =1$ if $y_i = y_j  $ and $0$ otherwise. In this case $\chi_k^{l,gt}= \dfrac{1}{N} \sum_i \dfrac{1}{k} \sum_j {A^l}_{ij} A^{gt}_{ij} $ is the average fraction of neighbors of a given point in $l$ that are in the same class as the central point (see Fig.~\ref{fig:cartoon}).
We set $k$ to one tenth of the number of images per class, but we verified that our findings are robust with respect to the choice of $k$ over a wide range of values (see Sec. \ref{sec:app_scaling}).
When calculated using the ground truth adjacency matrix as a reference,  $\chi_k^{l,gt}$ reduces  to the neighboring hit \cite{NH}. 
A measure of overlap quantitatively similar to $\chi_k^{l,m}$ can be obtained by using the method in \cite{cka} with a gaussian kernel of very small width (see Sec. \ref{sec:app_cka}). 

\subsection{Estimating the probability density}\label{sec:density_peak}
We  analyse the structure of the  probability density of data representations following the approach in \cite{cluster}, which allows to find the peaks of the data probability distribution  and the location and the height of the saddle points between them.
This in turn provides information on the relative hierarchical arrangement of the probability peaks.

The methodology works as follows.
Using a kNN estimator the local volume density  $\rho_i$ around each point $i$ is estimated.
The maxima of $\rho_i$ (namely the probability peaks) are then found.
Data point $i$ is a maximum if the following two properties hold: (I) $\rho_i>\rho_j$ for all the points $j$ belonging to $\mathcal{N}_k(i)$; (II) $i$ does not belong to the neighborhood $\mathcal{N}_k(j)$  of any other point of higher density \cite{cluster}.
A different integer label $\mathcal{C} = \{c^1, ... c^n\}$ is assigned to each of the $n$ maxima, and the data points that are not maxima are iteratively linked to one of these labels, by assigning to each point the same label of its nearest neighbor of higher density.
The set of points with the same label corresponds to a \emph{probability peak}.

The saddle points between probability peaks are then found.
A point  ${\bf x}_i \in c^\alpha$ is assumed to belong to the border $\partial_{c^\alpha, c^\beta}$ with a different peak $c^\beta$ if it exists  a point ${\bf x}_j \in \mathcal{N}_k(i) \cap c^\beta$ whose distance from $i$ is smaller than the distance from any other point belonging to  $c^\alpha$.
The saddle point between $c^\alpha$ and $c^\beta$ is the point of maximum density in $\partial_{c^\alpha, c^\beta}$.

Finally, the statistical reliability of the peaks is assessed  as follows.
Let $\rho^\alpha $ be the maximum density of peak $\alpha$, and $\rho^{\alpha,\beta} $ the density of the saddle point between $\alpha$ and $\beta$.
If $\log \rho^\alpha - \log \rho^{\alpha,\beta} < 2 Z \sqrt{  (4k+2)/[k(k+1)] } $, peak $\alpha$ is merged with peak $\beta$ since the value of its density peak is considered indistinguishable from the saddle point at a statistical confidence fixed by the parameter $Z$ \cite{cluster}.
The process is repeated until all the peaks satisfy this criterion, and are therefore statistically robust with a confidence $Z$.

\subsection{The dataset and the network architecture}\label{sec:data}
We  perform our analysis on the ILSVRC2012 dataset, a subset of $1000$ mutually exclusive classes of ImageNet which can be considered leaves of a hierarchical structure with 860 internal nodes.
The highest level of the hierarchy contains seven classes but $95\%$ of the ILSVRC2012 images belong to only two of these (artifacts or animals) and are split almost evenly between them ($55\%$ and $45\%$ respectively). 
Unless otherwise stated, the analysis in this work is performed on a subset of $300$ randomly chosen categories, including $300$ images for each category, for a total of $90,000$ images.

We extracted the activations of the training set of ILSVRC2012 from a selection of PyTorch pre-trained ResNet \cite{resnet} and VGG  \cite{vgg} networks. 
We measure our quantities on the output of each ResNet block and on the pooling and final fully connected layers of VGGs (checkpoints). These are the layers where  all the architectures downsample the channels (see Sec. \ref{sec:app_nets}) and the learned representations become more abstract and invariant to details of the input irrelevant for the classification task \cite{bengio2013representation}. This allows making a direct comparison between VGGs and ResNets architectures of different depths.

\paragraph{Reproducibility}  We include the source code of our experiments with the instructions required to run it on a selection layers in the supplementary material.

\section{Results}
It is well known that neural networks modify the representations of the data from an initial state where all the data are randomly mixed, to a final state where they are orderly clustered according to their ground truth labels \cite{frosst2019analyzing}. 
But where in the network does this order arise? In the output layer the nearest neighbors of, say, the image of a cat are very likely other images of cats. But in which layer, and in which manner do cat-like images come together?
We describe the ordering process by analyzing the change in the probability distribution across the layers. 

\subsection{The evolution of the neighbor composition in a DCN}\label{sec:nucl}
\begin{figure}[!tp]
\vspace{-0.5cm}
\centering
\includegraphics[width=0.84\columnwidth]{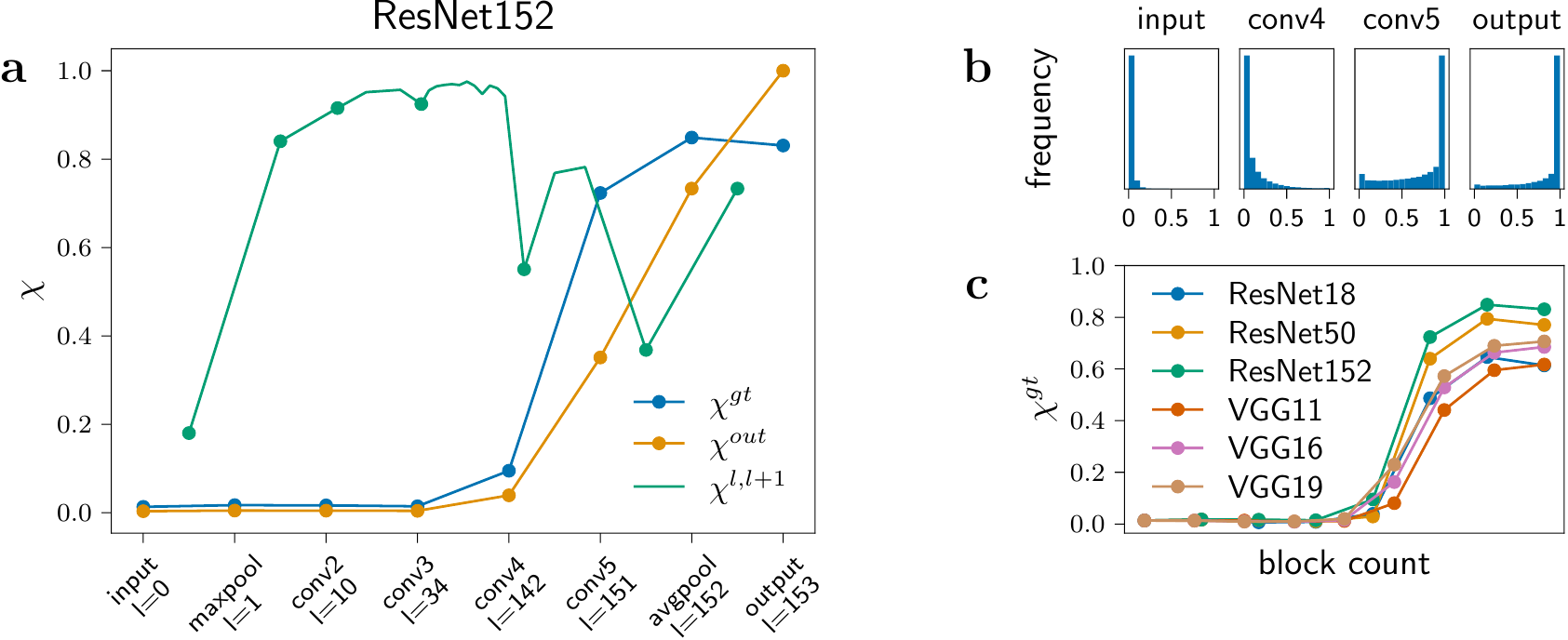} 
\caption{ {\bf Overlap profiles in ResNet152 and in different architectures}. ({\bf a}:) Overlap between the checkpoint layers and the ground truth $\chi^{l,gt}$ (blue) and with the output $\chi^{l,out}$ (orange). The green profile shows the overlap between nearby layers $\chi^{l,l+1}$ with dots in correspondence to the checkpoints. 
({\bf b}:) Probability distribution of $\chi^{l,gt}$ for four layers.
({\bf c}:) Profiles of $\chi^{l,gt}$ for six architectures of different depths. The values of $\chi^{l,gt}$ measured on the checkpoints are displayed uniformly on the $x$-axis. 
}
\label{fig:ov_gt}
\vspace{-0.5cm}
\end{figure}

We first characterize the probability flux  by computing the neighborhood overlap $\chi^{l,out}_k$: the fraction of $k$-neighbors of a data point which are the same in layer $l$ and in the output layer (Eq. \ref{eq:overlap}). 
Figure \ref{fig:ov_gt}-a shows the  behaviour of  $\chi_{k}^{l,out}$ as a function of $l$ for the checkpoint layers of the ResNet152 described in Sec. \ref{sec:data} (orange line).
The neighborhood overlap remains close to zero up to $l$=142.
In the next 9 layers it starts growing significantly, reaching a value of 0.35 in layer 151 and 0.73 in layer 152, the last before the output.
In the same figure, we also plot the neighborhood overlap of each layer with the ground truth classification $\chi^{l,gt}_k$ (blue line).
After layer 142, $\chi^{l,gt}_k$ changes even more abruptly than $\chi^{l,out}_k$, increasing from 0.10 to 0.72. 

We can obtain more insights into this transition process looking at the probability distribution of $\chi^{l,gt}$ across the dataset in four different layers (see Fig. \ref{fig:ov_gt}-b). 
In the input and output layers the probability distribution is unimodal.
In layer 142 (conv4), before the onset of the transition, the distribution is still strongly dominated by disordered neighbors, but an ordered tail starts to emerge. 
In layer 151 (conv5), immediately after the transition, the distribution indicates the coexistence of some data points in which the neighborhood is still disordered or only partially ordered ($\chi \approx 0$) and some data in which it is already ordered ($\chi \approx 1$).

These results show that ordering, when measured by the consistency of the neighborhood of data points with respect to their class labels, changes abruptly, in a manner which qualitatively resembles the phase transition of a \q{nucleation} process. 
The green profile of \ref{fig:ov_gt}-a reinforces this evidence showing the overlap between two nearby layers $\chi^{l, l+1}$. This quantity is a measure of the probability flux between any two consecutive layers. In the first layer the neighborhoods are almost completely reshuffled, as indicated by an $ \chi^{0, 1} \sim 0.2$. Afterwards, up to layer 142 $ \chi^{l, l+1} \sim 0.95$, indicating that the neighborhoods change their compositions smoothly like in a slow diffusion process. In the two central blocks, from layer 10 to layer 142, it takes 20-30 layers to change half of the neighbors of each data point, i.e. to decrease $\chi^{l, l'}$ to 0.5 (see Sec. \ref{sec:app_memory}).
At layer 142 instead, the first ordered nuclei appear and $\chi^{l, l'}$ drops to 0.55 in just one layer.
A significant reshuffling of the neighborhoods takes place at layer 151, where $\chi^{l, l+1}$ drops again to 0.61.
We will see in Sec. \ref{sec:cluster} that  in  this layer the structure of the probability density changes significantly, and the probability peaks corresponding to the "correct" categories appear. 
The effective \q{attractive force} acting between data with the same ground truth label overcomes the entropic-like component coming from the intrinsic complexity of the images, and clusters of akin images emerge almost all at the same time (i.e., at the same layer), giving rise to a sharp transition.

Is the sudden change we observed specific to this architecture or is it a common feature of deep networks trained for similar tasks? 
To answer this question we repeated the same experiment on architectures of different sizes of the ResNet and the VGG families.
We observed a common trend in all the architectures, as depicted in Fig.~\ref{fig:ov_gt}-c, where we plot $\chi^{l,gt}$ for the checkpoints described in Sec. \ref{sec:data}.
In all the cases, $\chi^{l,gt}$ remains close to zero for many layers, and then sharply increases in only a few layers towards the end of the network.
The value of $\chi^{l,gt}$ in the output layer is different in different architectures, consistently with the fact that their classification accuracy is different. 
\subsection{The data landscape before the onset of ordering}\label{sec:entropy}

\begin{wrapfigure}{r}{0.38\linewidth}
\vspace{-0.cm}
\includegraphics[width=0.38\columnwidth]{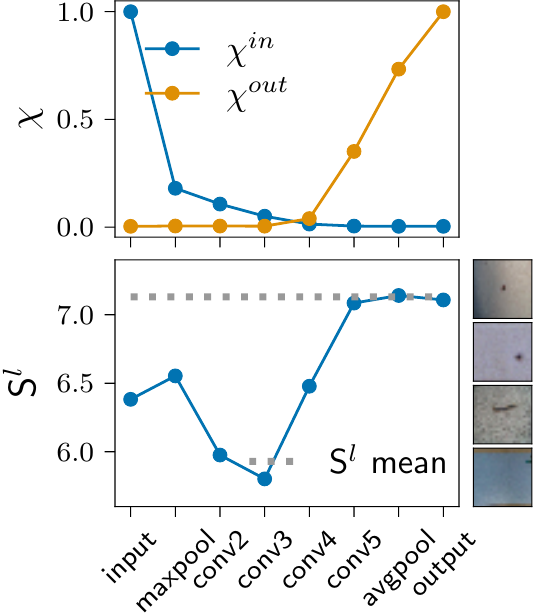}
\caption{{\bf  Image entropy in ResNet152.} ({\bf Top}:) Overlap with the input $\chi^{in}$ (blue), and output $\chi^{out}$ (orange) layers.
({\bf Bottom}:) Average image image entropy $S^l$ within the first 30 neighbors; the errorbars are shorter than the marker size; the most frequent images found in conv3 are shown on the right.}
\label{fig:2_ov_ll_entr}
\vspace{-1.0cm}
\end{wrapfigure}

It has been argued that the first layers of deep networks serve the important task of getting rid of unimportant structures present in the dataset \cite{shwartz2017opening, achille2018emergence, id, lecun2015deep}.
This phenomenon is illustrated in the upper panel of Fig. ~\ref{fig:2_ov_ll_entr}, which  shows that any overlap with the input layer is lost roughly after the conv3 landmark (layer 34). 

We found that in intermediate layers  the DCNs analysed in this work construct high-dimensional hyperspherical arrangements of points with very few \q{simple} images at the center. This is related to the high intrinsic dimension (ID) of these layers \cite{id}.
When the ID is very high, few data points act as \q{hubs}\cite{hubs}, namely they  fall in a large fraction of the  other point's
 neighborhoods while the others fall in just a few.
Moving from the input to conv3 the same images appear in a growing number of neighborhoods.
In conv3 the top ten most frequent images are found in almost half of the 90,000 neighborhoods with a high of 75,663 for the most frequent of all.

Hub images  are particularly \q{simple}, looking in most cases like elementary patterns (dots, blobs, etc.) lying on almost uniform backgrounds (see Fig. ~\ref{fig:2_ov_ll_entr}, bottom right).
To quantify this perceptual judgment we computed the Shannon entropy of an image $S = -\sum_{n_c}\sum_{v} p_v  \log_{2}(p_v)/n_c$ where $p_v$ is the normalized frequency of pixels of value $v$ and $n_c = 3$ is the number of channels of RGB images. The average entropy of the neighbors of an image $i$ in a layer $l$ is given by $S_i^l = \sum_{j \in \mathcal{N}_k(i)} S_j^l /k$, and averaging across all images we obtain a  measure of the neighborhood entropy of a layer $S^{l}$. A low value of $S^{l}$  means that, in that layer, the neighborhoods contain many low-entropy images. 
In bottom panel of Fig. ~\ref{fig:2_ov_ll_entr} we show how  $S^{l}$ changes: in intermediate layers, and most prominently in conv3 -- where the intrinsic dimension is at its peak (see Sec. \ref{sec:app_modmnist}) -- the representation is organized around low-entropy hubs whose centers are low-$S$ images (blue line, and left stack of hub images).
As a reference, we also report the entropy computed shuffling the neighbors assignments (grey dashed line).
\subsection{The evolution of the probability density across the hidden layers}\label{sec:cluster}

We have seen that at layer34 (conv3)  all images are arranged in a high-dimensional hypersphere  and that at layer 151 (conv5) the neighborhoods are already organized consistently with the classification.    
Clearly, the most important transformations of the representation happen in between these layers. 
To shed some light on the evolution of the representations in these intermediate layers we hence use a tool which allows characterizing multidimensional probability distributions, finding its probability peaks and localizing all the saddle points separating these peaks (see Sec. \ref{sec:density_peak}).
We will see that the nucleation-like transition described in Sec. \ref{sec:nucl} is a complex process, in which the network separates the data in a gradually increasing number of density peaks laid out in a hierarchical fashion that closely mirrors the hyperonymous-hyponymous relations of the ILSVRC2012 dataset.
\begin{figure}[!b]
\vspace{-0.4cm}
    \centering
\includegraphics[width=0.96\columnwidth]{{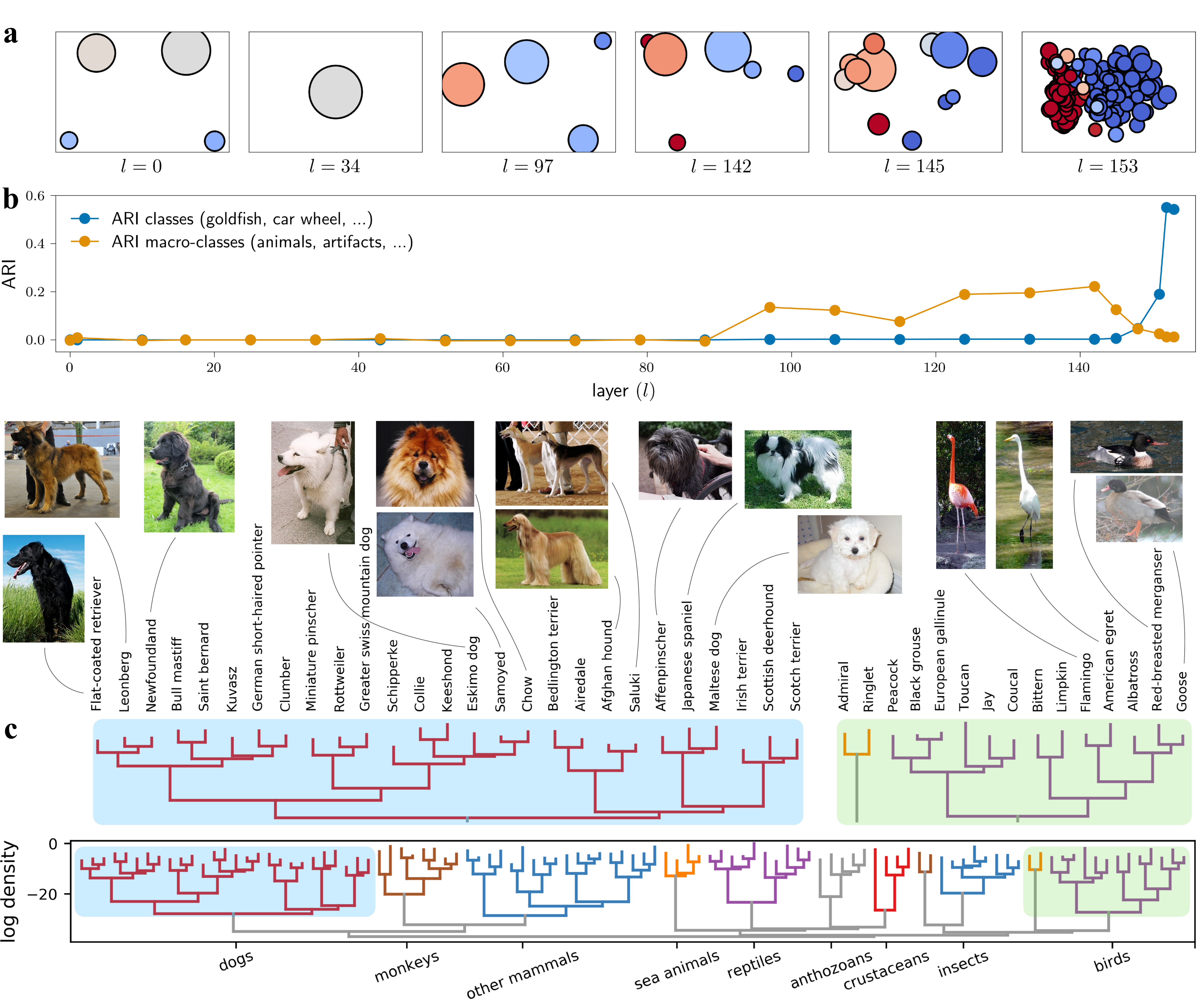}}
    \caption{{
    \bf Structure and composition of density peaks of representations}. 
    ({\bf a}:) A schematic view of the peaks in 6 layers. Color tones refer to the relative presence of animals and artifacts in each peak: dark red = $100\%$ of animals, dark blue = $100\%$ of artifacts.
    ({\bf b}:) $ARI$ profiles for animal/artifact partition and the 300 low-level classes (blue and orange). 
    ({\bf c}:) The dendrogram portrays the hierarchical connections between the density peaks of the animal branch. On the $y$-axis the value of the density peaks is plotted in logarithmic scale. Two insets above show the detailed composition of specific branches (light blue and light green). }
    \label{fig:dendogram}
\end{figure} 

Figure~\ref{fig:dendogram}-a shows a two-dimensional visualization of the number and organization of the probability peaks of the representations in some of the layers.
In the input layer ($l$=0), the data are split into two major peaks, which roughly divide the training set into light and dark images. 
This structure is not useful for classification, and is wiped out within the first 34 layers of the network. In conv3 the probability density becomes unimodal, consistently with the analysis of the previous section.
In the subsequent layers the network creates structure that is useful for classification, and in layer 97 a bimodal distribution appears. The other peaks shown in figure are very small and retain only a few hundreds data points each.
The same density peaks persist until layer 142, where $97\%$ of the images still reside in the two biggest ones.
Finally, after layer 142 the two large peaks break down into smaller ones representing individual classes.

To asses the population of the density peaks in terms of ground-truth categories we use the Adjusted Rand Index ($ARI$) \citep{rand_indx,ref-rand_indx}. 
Roughly speaking, $ARI$ is zero if the density peaks do not correspond to the reference partitions of the dataset, and is one if they match it. 
In Fig. \ref{fig:dendogram}-b we plot the $ARI$ with respect to the high level animal/artifact categories ($ARI^{macro}$, orange line), and with respect to the $300$ low level classes we sampled ($ARI^{cl}$, blu line). 
From layer 97 to layer 142 artifacts and animals predominantly populate one of the two major peaks increasing the $ARI^{macro}$ value to 0.22 while the correlation with low level classes remains absent.
The following breakup of the peaks leads to a drop of $ARI^{macro}$ to 0 and to a concomitant sharp rise of $ARI^{cl}$ from 0 to 0.55, consistent with the nucleation mechanism detected by $\chi^{l, gt}$ and described in Sec. \ref{sec:nucl}. 
Moreover, some classes are separated before others (Fig.~\ref{fig:dendogram}-a, layer 145), consistently with the bimodality in the distribution of $\chi^{gt}$ observed in the bottom panels of Fig.~\ref{fig:ov_gt}-a.
Interestingly, many of the density peaks in the layers between 142 and 153 (i.e., during the nucleation transition) closely resemble the hierarchical structure of the concepts in ILSVRC2012.
For instance, in layer 148 one can find peaks corresponding to insects, birds, but also ships and buildings (see Sec. \ref{sec:app_peaks}).

In the last layer ($l$ = 153) the different peaks  correspond to the different classes, but the structure of the probability density is much richer than a simple collection of disjointed peaks. Indeed  the hierarchical process that shaped the density landscape across the layers leaves a footprint on the organization of the peaks.
For instance, the division in macro-classes of animals and artifacts formed in layer 97 is still present in the last layer as indicated by the fact that red and blue clusters are found primarily on the left and on the right of the corresponding low dimensional embedding (Fig.~\ref{fig:dendogram}-a).
But much more structure is present.
In Figure \ref{fig:dendogram}-c we visualize the probability landscape of the animal classes as a dendrogram, in which each leaf corresponds to a peak, and the leaves are merged sequentially, following the WPGMA algorithm \cite{sokal1958statistical}, according to the height of
the saddle point of the probability density between them.
In this manner the secondary probability peaks belonging to the same large scale structure form a branch of the dendrogram. The height of a leaf in Fig. \ref{fig:dendogram}-c is proportional to  the logarithm of the density of the peaks.
The morphological similarities of animals with similar genetic material make it possible for the dendrogram in Fig. ~\ref{fig:dendogram} to reproduce the taxonomy of a phylogenetic tree to an astonishing degree.
At the root of the dendrogram, we can notice a first distinction between mammals on the left and other animals on the right.
At the following herarchical level we can find a more specific separation of animal types.
Dogs, reptiles, birds and insects and so on can be easily identified.
Finally, within each species, say dogs (Fig. \ref{fig:dendogram}), alike breeds are linked by tighter bounds, that is saddle points of higher density.

In the supplementary material we include the values of the probability density and the integer identifier of the density peak to which each image belongs for the relevant representations analysed in this section.
We also provide the topography of the probability density, namely the height of all the  peaks and of all the saddle points between them.

\section{Discussion and Conclusion}\label{sec:conclusion}

This paper presents an explicit characterization of the evolution of the probability density of the data landscape across the layers of a DCN.
We showed that this probability density undergoes a sequence of transformations which brings to the emergence of a rugged and complex probability landscape. 
Rather surprisingly, we found that the development of these structures is not gradual, as one would expect in a deep network with more than one hundred layers.
Instead, the greatest changes to the neighborhood composition and the emergence of the probability peaks are localized in a few layers.
This picture seems qualitatively different with the one emerging from SVCCA \cite{svcca}, projection weighted CCA  \cite{morcos2018insights} and linear CKA \cite{cka}, which have revealed smoother changes between nearby representations. 
A first reason of this difference lies in the kind of correlation captured by these similarity indices. 
The ordering mechanism starting with the separation between animals and artifacts is functional and correlated to a successful fine-grained classification of the categories. In essence CCA based methods capture the correlation between the final categories (the peaks) and the \q{intermediate level} concepts (\q{the mountain chains}) required to construct them which are recognized already in the middle layers of the network.
The overlap defined in Eq. \ref{eq:overlap} measures instead a correlation growing only when the neighbors become consistent with those of the output ($\chi^{l,out}$) or their labels ($\chi^{l,gt}$). In section \ref{sec:app_cka} of the appendix we show that $\chi^{l,out}$ is similar to the correlation obtained by Gaussian CKA\cite{cka} using a very small kernel width.

A second possible reason for the discrepancy between the results reported in this work and those reported in the literature is the complexity of the datasets analysed. Indeed, most previous studies have focused on datasets like MNIST and CIFAR-10.
These datasets lack the semantic stratification of ImageNet and hence show a much smoother evolution of the probability landscape, because in these datasets the number steps needed to disentangle the  hierarchy of features of the categories is smaller (see Sec. \ref{sec:app_modmnist}).
We are unaware of attempts that directly targeted the similarity of DCN representations in connection to the hierarchical structure of a complex dataset like ImageNet.
In \cite{bilal2017convolutional}, confusion matrices have been used to visually analyse the correlations between classes showing results in agreement with our conclusions.
However, the algorithm we use here (Sec. \ref{sec:density_peak}) relies on density estimation 
and is able to reconstruct a probability landscape that faithfully follows the hierarchical structure of categories (Sec. \ref{sec:cluster}) in an unsupervised manner, with no need to consider the ground truth labels and estimating the confusion matrix; indeed, our approach works also in the limiting case of 100\% test accuracy.

We believe that the detailed picture of the evolution of the probability density provided in this work can trigger a more rational design of the architecture and of the learning protocols of DCNs. One can imagine to define training losses targeting the development of probability peaks according to a pre-defined semantic classification. This can be enforced  in the intermediate layers of a network, rather than only in the output layer, somehow enhancing the separation between macro categories arising spontaneously for this dataset. 
One can even imagine to use the topography of the density peaks developed by a deep neural network as a hierarchical classifier, going well beyond the sharp classification in categories.  An appropriate understanding of the nucleation mechanism could also be beneficial to transfer learning, since it gives a simple rational criterium to judge the generality (i.e. transferability) of the features of a representation \cite{yosinski2014transferable}.
Turning to synergies between artificial neural networks and neuroscience \cite{barrett2019analyzing},  
 we forsee many potential uses of this methodology among which 1) assessing the local similarity of cortical representations in different brain areas 2) localizing the cortical areas that code more explicitly for a certain set of features and 3) monitoring how neighbor relationships change during learning. 

\section*{Broader impact}
This work does not present any foreseeable societal consequence.

\newpage

\bibliographystyle{unsrtnat}
\bibliography{references}

\newpage
\appendix

\renewcommand\thefigure{\thesection.\arabic{figure}}  
  
\section{Appendix}
\subsection{Details on the architectures considered}\label{sec:app_nets}

We briefly describe here the structure of the VGG \cite{vgg} and ResNet \cite{resnet} architectures analysed in our work. 
The first part of the architectures is composed of convolutional and pooling layers. A convolutional layer maps a stack of $w$ features (or channels) of size $l_{in} \times l_{in}$ into another through a filter.
In ResNets and VGGs the size of the filter is mostly $3 \times 3$, their width is always equal to $w$. The result of a convolution is then passed through a ReLU and produces one output channel. Different filters produce different output channels. When the size $l_{out}$ of the output channel is halved the number of filters is doubled. In VGGs channels are downsampled by pooling layers, in ResNets mainly doubling the filter stride.
Finally, VGGs end with three fully connected layers, ResNets with only one.
In our study we used the convolutional layers that downsample the channels together with the fully connected layers as chackpoints.

\setcounter{figure}{0} 
\subsection{Scaling of \texorpdfstring{$\chi^{l,gt}$}{cgt} and \texorpdfstring{$\chi^{l,out}$}{cout}}\label{sec:app_scaling}
\begin{figure}[H]
\vspace{0cm}
\centering
{\includegraphics[width=0.9\columnwidth]{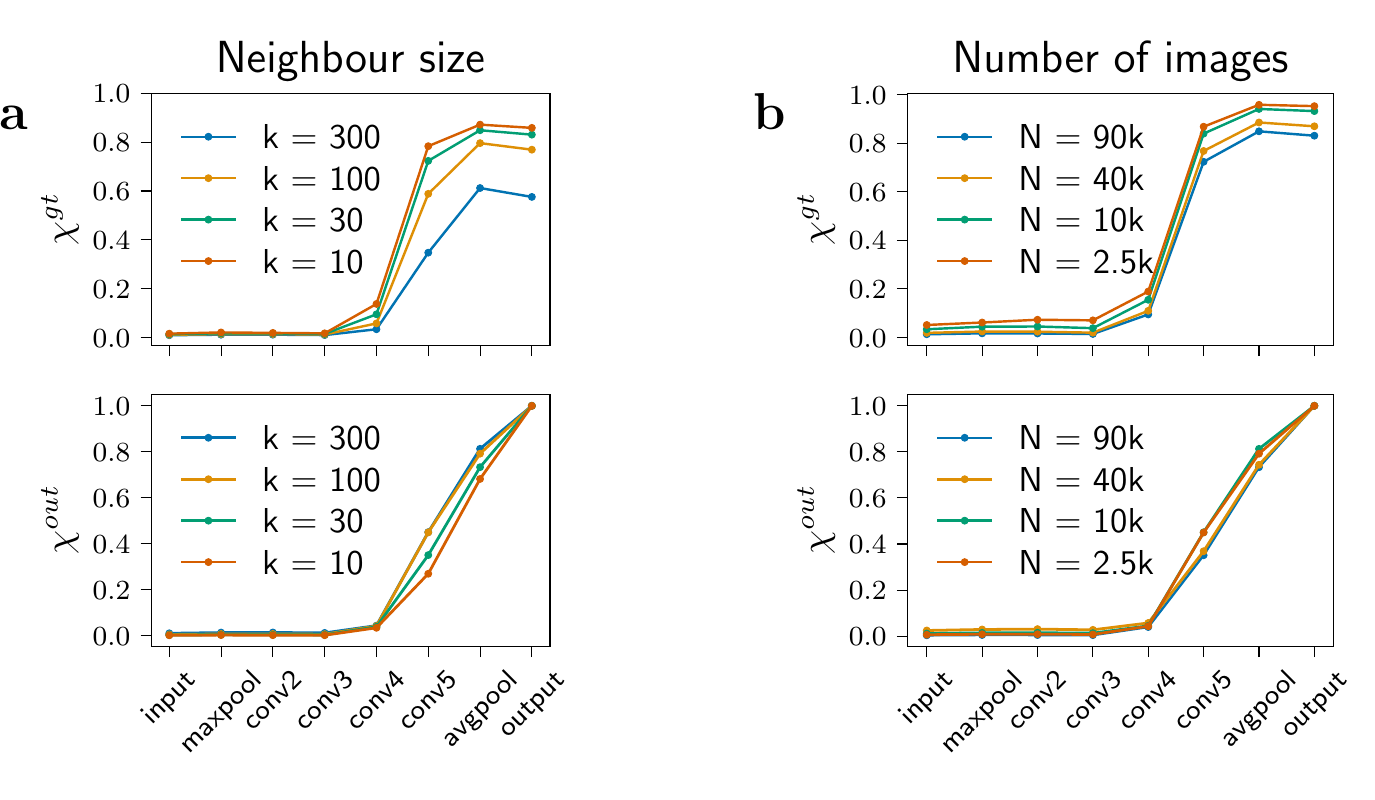}}
	\caption{({\bf a}:) Profiles of the overlap with the ground truth labels (top) and with the output layer (bottom) as a function of the neighbor size. ({\bf b}:) Profiles of the overlap with the ground truth labels (top) and with the output layer (bottom) as a function of the total number $N$ of images.}
\label{fig:app1}
\end{figure}

Figure \ref{fig:app1}-a shows the overlap with the ground truth $\chi^{l, gt}$ (top) and with the output activations $\chi^{l, out}$ (bottom) in ResNet152, for the same subset of 90,000 examples from ILSVRC2012 analysed in  Sec. \ref{sec:data}. In our experiments we empirically set $k=30$ i.e. one tenth of the number of images per class.  Figure \ref{fig:app1}-a shows that the trend of $\chi^{l, gt}$ and $\chi^{l, out}$ is rather robust over a wide range of $k$-values. Only when  $k$ is very large ($k=300$)   the transition in the last layers of the network is not detected very clearly.

In Figure \ref{fig:app1}-b we plot how $\chi^{l, gt}$ (top) and $\chi^{l, out}$ (bottom) vary with the dataset size $N$. As the number of examples $N$ increases we keep the ratio between the number of classes and images per class constant.  This  shows that the results are also robust with respect to $N$.

\subsection{Overlap with the checkpoint layers}\label{sec:app_memory}
\begin{figure}[H]
\centering
\vspace{0.cm}
\includegraphics[width=0.5\columnwidth]{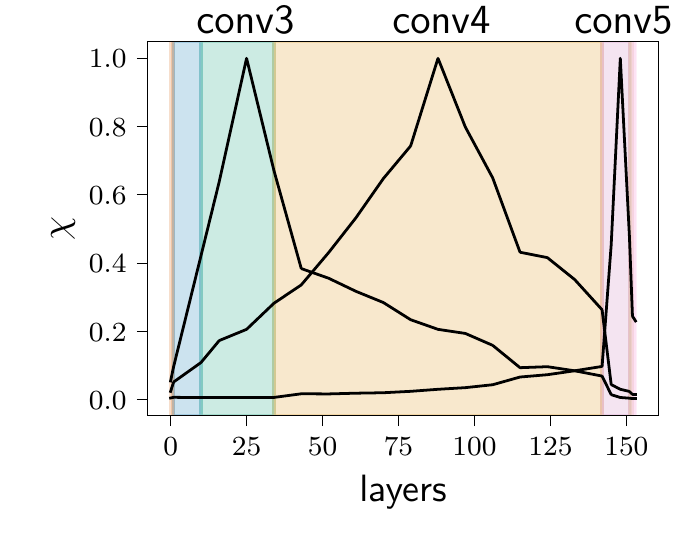}
\caption{Overlap with layers 25, 88, 148 in ResNet152. Different background colors indicate different ResNet blocks }
\label{fig:app3_memory}
\vspace{0cm}
\end{figure}

Figure \ref{fig:app3_memory} shows the overlap  of the representations with respect to the representation at  tree layers $l = 25$, $l = 88$ and $l= 148$, belonging to tree distinct ResNet152 blocks. 
On average the number of layers required to change half of the neighbors is $\sim 20$ in conv3 and $\sim 30$ in conv4, while in conv5 where the nucleation takes place the same change occurs in just one layer.
Indeed, the rate at which neighbors are reshuffled grows dramatically when the  ordered clusters appear (see Sec. \ref{sec:ov}). 
The neighborhood composition changes significantly also between two blocks when the channels are downsampled. 

\subsection{Central kernel alignment vs overlap}
\label{sec:app_cka}
\begin{figure}[H]
\centering
{\includegraphics[width=0.9\columnwidth]{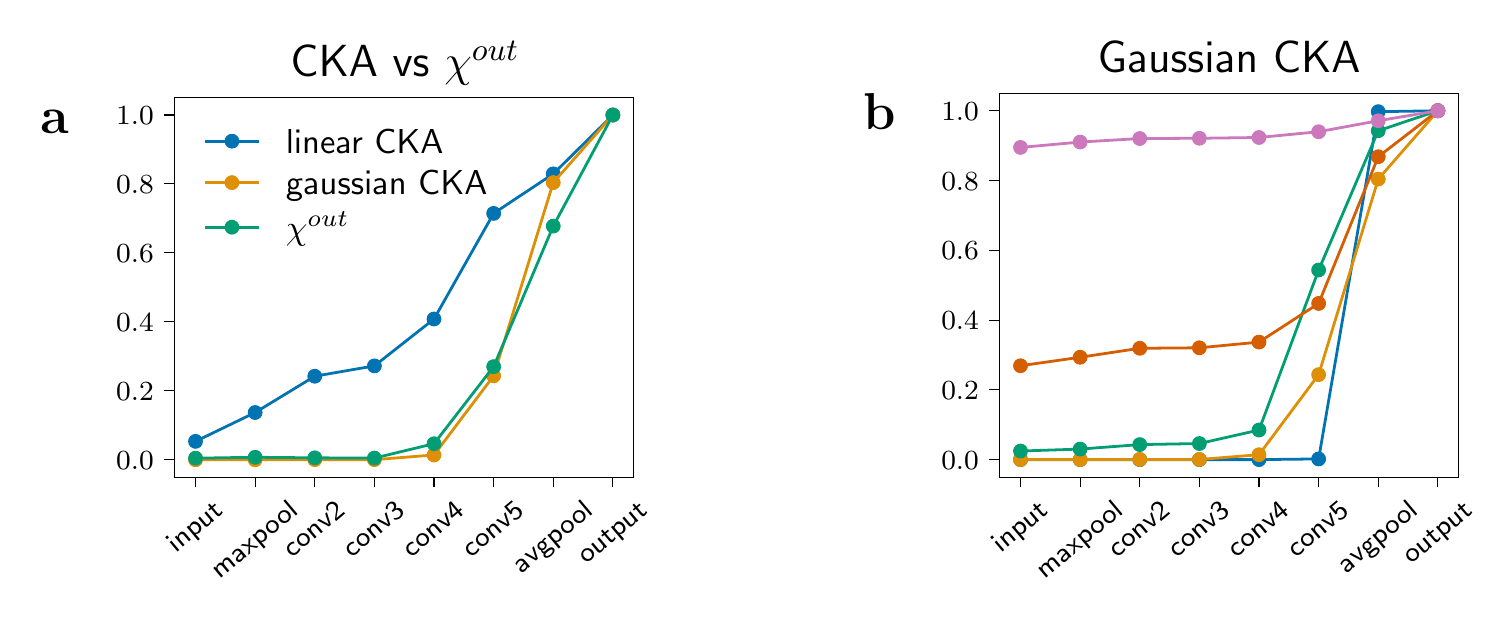}}
	\caption{({\bf a}:) Linear CKA (blu), overlap $\chi^{l, out}$ (green) and gaussian CKA (orange) with the output layer in ResNet152 for a subset of 5000 ILSVRC2012 images. We kept 50 classes and 100 images per class and set the kernel bandwidth $\sigma$ to 0.2 times the average distance with the first nearest neighbor $\overline{d_1}$ . ({\bf b}:) Gaussian CKA with the output layer as a function of the kernel badwidth $\sigma$: $\sigma = 0.1\overline{d_1}$ (blu),  $\sigma = 0.2\overline{d_1}$ (orange),  $\sigma = 0.5\overline{d_1}$ (green),  $\sigma = \overline{d_1}$ (red),  $\sigma = 2\overline{d_1}$ (pink).}
\label{fig:app_cka}
\end{figure}

Central kernel alignment (CKA) is the normalized squared Hilbert-Schmidt norm of the cross covariance operator between representations \cite{cka}. 
Like the neighborhood overlap it is invariant under orthogonal transformations and isotropic scaling but not to an arbitrary invertible linear transformation. This has been argued to be too a limitation for a similarity index between representations \cite{cka}.  Gaussian CKA probes the local similarity between representations and can seen as a kernel smoothing of the neighborhood overlap presented in Sec \ref{sec:ov}.  
In figure \ref{fig:app_cka}-a we show the gaussian CKA (orange) and the overlap (green) with the output layer setting the kernel bandwidth to 0.2 times the average distance with the first nearest neighbor. 
Linear CKA is equivalent to a CCA between representations  in which the canonical variates are weighted by the corresponding eigenvalues \cite{cka}. Linear CKA steadily increases already  in the early layers of the network (see Fig. \ref{fig:app_cka}-a blu profile).

Figure \ref{fig:app_cka}-b shows how the gaussian CKA with the output is affected by different choices of the kernel bandwidth $\sigma$. The smaller is $\sigma$ the sharper is the transition measured by the index.

\subsection{Overlap and intrinsic dimension profiles in different datasets}\label{sec:app_modmnist}
\begin{wrapfigure}{r}{0.44\columnwidth}
\vspace{-0.7cm}
{\includegraphics[width=0.4\columnwidth]{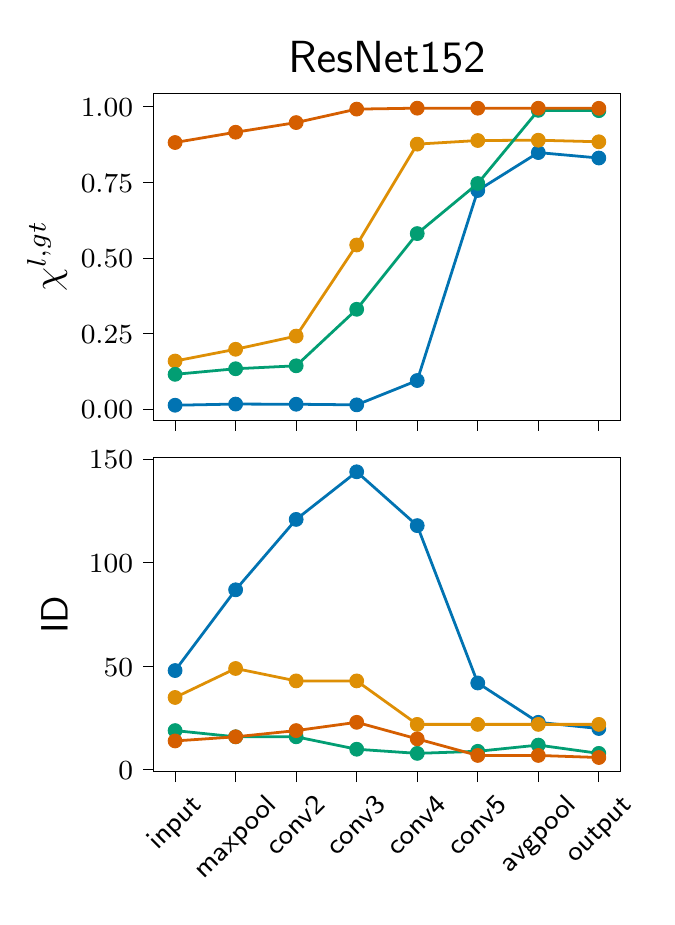}}
\caption{Overlap with the ground truth labels ({Top})  and intrinsic dimension profiles ({Bottom}) in ResNet152 for different datasets: MNIST (red), CIFAR10 (orange), modMNIST (green), ImageNet (blue).}
\label{fig:app3}
\vspace{-0.5cm}
\end{wrapfigure}
In this section we compare the overlap with the ground truth labels $\chi^{l, gt}$ and the intrinsic dimension (ID) profiles of different dataset of inceasing complexity in ResNet152.

The top panel of figure \ref{fig:app3} shows $\chi^{l, gt}$ for a ResNet152 architecture trained on MNIST \cite{lecun2010mnist} modMNIST, CIFAR-10 \cite{cifar10} and ImageNet \cite{imagenet09}. To generate the modMNIST dataset we resize the dimension of the MNIST  digits by a factor ranging from 0.2 to 0.4 and moved them in a random location of the image. We finally scale up the size of the images to 224x224 pixels. We trained MNIST and modMNIST for 10 and 20 epochs respectively using Adam optimizer
\cite{kingma2014adam} with default parameters (lr=0,001, betas=(0,9; 0,999)); we trained CIFAR10 for 120 epochs with stochastic gradient descent with momentum (lr = 0.1, momentum = 0.9), decreasing the learning rate by a factor 10 after 60 epochs; we used the PyTorch pre-trained ResNet152 model for ImageNet.

MNIST can be directly classified with high accuracy with a $k$-NN estimator. Consistently already in the input layer  $\chi^{l,gt} \approx 0.78$ and reaches one from conv3 onwards.
In modMNIST and CIFAR-10 the categories are only 10, therefore the initial values of $\chi^{l, gt}$ are larger, the lag phase is shorter the one   of ImageNet.
While qualitatively, $\chi^{l, gt}$  behaves similarly in modMNIST, CIFAR-10 and ILSVRC2012, the transition of $\chi^{l, gt}$ seems to be sharp only for the ILSVRC2012 dataset, and is therefore likely related to the complexity of the prediction task. 

Bottom panel shows the intrinsic dimension (ID) for the same datasets across the checkpoints layers of ResNets152. The higher the complexity of the dataset the more are the factors of variations encoded in the embedding manifold, the higher is the ID. 
For complex datasets like ImageNet the ID has the hunchback shape reported in \cite{id}, while for MNIST and modMNIST it is almost constant, and it takes much smaller values, uncorrelated with $\chi^{l,gt}$ . This  supports the hypothesis that the  transition observed in the value of the neighborhood overlap is not necessarily related with a sharp change of the intrinsic dimension of the representation.

\subsection{Details of density peaks appearing between layer 142 and 148}\label{sec:app_peaks}

\begin{figure}
    \centering
    \includegraphics[width = \columnwidth]{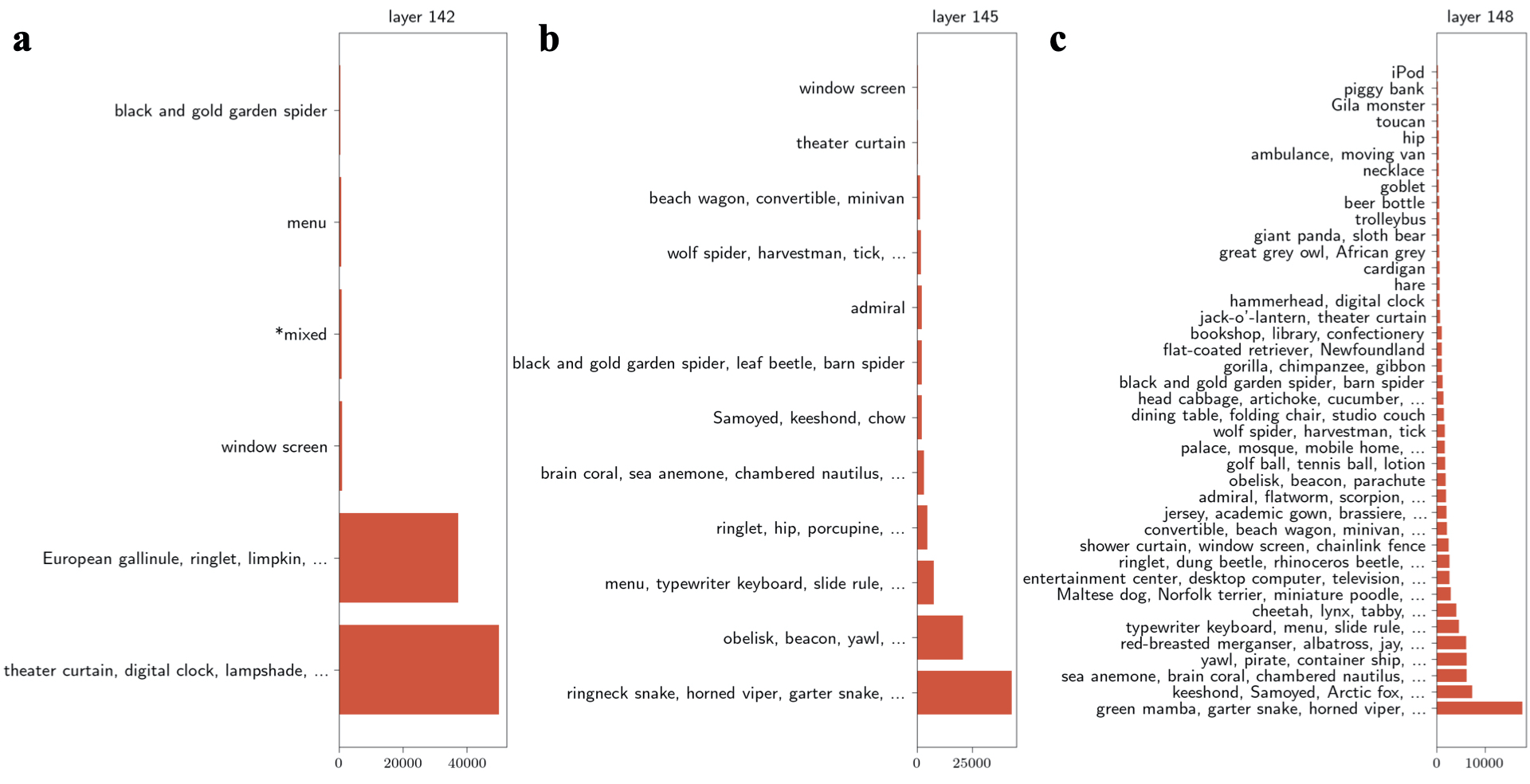}
    \caption{Composition of density peaks in layers 142 (\textbf{a}), 145 (\textbf{b}) and 148 (\textbf{c}). The x-axis indicates the size of the peak, the y-axis reports the categories represented with more than 150 points in the peak. Consecutive dots (\q{...}) indicate that more than three categories are well represented in the peak. The peaks are ordered from the smallest to the largest from top to bottom.}
    \label{fig:app_layers_142_148}
\end{figure}

In figure \ref{fig:app_layers_142_148} we report a visualisation of the the density peaks appearing during the \q{nucleation transition} of Resnet 152. 
In particular, the image shows the size and approximate composition of the  peaks present in the layers 142, 145, and 148.
As discussed in Section \ref{sec:cluster}, in layer 142 the data density is dominated by two large peaks composed of images of animals and artifacts respectively.
This structure is visible in panel (a), in which one can easily identify the two large peaks.
In the subsequent layers, the animal and artifact peaks break down into small peaks containing images of the same class.
The process happens in a hierarchical fashion: peaks corresponding to multiple classes sharing a lot of semantic similarities appear first, and subsequently break down into smaller peaks corresponding to the single classes.
This phenomenon can be observed in panels (b) and (c).
For instance, in layer 145 (panel (b)) one can clearly identify peaks corresponding to certain kinds of arachnids (wolf spider, harvestman, tick, ...), insects (black and gold spider, leaf beetle, barn spiders) 4-wheel means of transportation (beach wagon, convertible, minivan), dogs (Samoyed, keeshond, chow), and so on.
In layer 148 (panel (c)) this process continues and one finds many more peaks, corresponding either to single classes (e.g., iPod, piggy bank and beer bottle) or to groups of similar classes.
At the end of the nucleation process described, from layer 152 (not shown here) one finds approximately one peak corresponding to each class label.

\end{document}